%% file: main.tex
\newcommand{\sys}{InstAgent}
\newcommand{\projectpageurl}{https://instagent01.github.io/}

\newcommand{\papertitle}{%
  \sys: A Recursively Self-Improving Agent Operating System
  for Heterogeneous Robot Collaboration%
}
\newcommand{\memname}{SMAR-T}

\newcommand{\paperauthors}{%
  Haoyang Luo, Junhao Qin, Jingyan Cao,
  Yi Huang, Guangjie Yang, Xiaoyu Chen, Haotao Lu,
  Tianxiang Gui, Ye Shi, Jingya Wang%
}

\newif\ifanonymous

\anonymousfalse

\documentclass[letterpaper,10pt,conference]{ieeeconf}

\IEEEoverridecommandlockouts
\usepackage[T1]{fontenc}
\usepackage{mathptmx}
\usepackage{amsmath,amssymb}
\usepackage{graphicx,booktabs,array,tabularx}
\usepackage{xcolor,tikz}
\usepackage{pgfplots}
\pgfplotsset{compat=1.18}
\usepgfplotslibrary{groupplots}
\usetikzlibrary{arrows.meta,positioning,fit,calc}
\usepackage{algorithm,algpseudocode}
\usepackage{multirow} 

\makeatletter
\newcommand{\fs@icraruled}{%
  \fs@ruled
  \def\@fs@pre{%
    \vskip6pt
    \hrule height.8pt depth0pt
    \kern2pt
  }%
}
\makeatother

\floatstyle{icraruled}
\restylefloat{algorithm}

\usepackage{cite,url}

\usepackage[hidelinks]{hyperref}

\ifanonymous
  \hypersetup{pdfauthor={},pdftitle={\papertitle}}
\else
  \hypersetup{pdfauthor={\paperauthors},pdftitle={\papertitle}}
\fi

\definecolor{draftred}{RGB}{158,45,40}
\definecolor{agentblue}{RGB}{37,76,115}
\definecolor{memorygreen}{RGB}{43,105,86}

\title{\LARGE\bf \papertitle}

\ifanonymous
  \author{}
\else
    \author{%
      Haoyang Luo$^{1,2,*}$,
      Junhao Qin$^{1,2,*}$,
      Jingyan Cao$^{1,2*}$,
      Yi Huang$^{1}$,\\
      Guangjie Yang$^{1}$,
      Xiaoyu Chen$^{1}$,
      Haotao Lu$^{1,2}$,\\
      Tianxiang Gui$^{1,2}$,
      Ye Shi$^{1,2}$,
      and Jingya Wang$^{1,2,\dagger}$%
      \thanks{%
        $^{*}$Haoyang Luo, Junhao Qin, and Jingyan Cao
        contributed equally to this work.%
      }%
      \thanks{%
        $^{1}$ShanghaiTech University, Shanghai, China.%
      }%
      \thanks{%
        $^{2}$InstAdapt, Shanghai, China.%
      }%
      \thanks{%
        $^{\dagger}$Corresponding author: Jingya Wang.%
      }%
    }
\fi

\date{}

\IEEEaftertitletext{\input{figures/00_fig1}}

\begin{document}

\maketitle
\thispagestyle{empty}
\pagestyle{empty}

\input{sections/00_abstract}
\input{sections/01_introduction}
\input{sections/02_related_work}
\input{sections/03_method}
\input{sections/04_experiments}
\input{sections/05_conclusion}
\input{sections/06_acknowledgments}

\bibliographystyle{IEEEtran}
\bibliography{references}

\end{document}

%% file: figures/00_fig1.tex
\begin{minipage}{\textwidth}
\makeatletter\def\@captype{figure}\makeatother
\centering
\includegraphics[width=\linewidth]{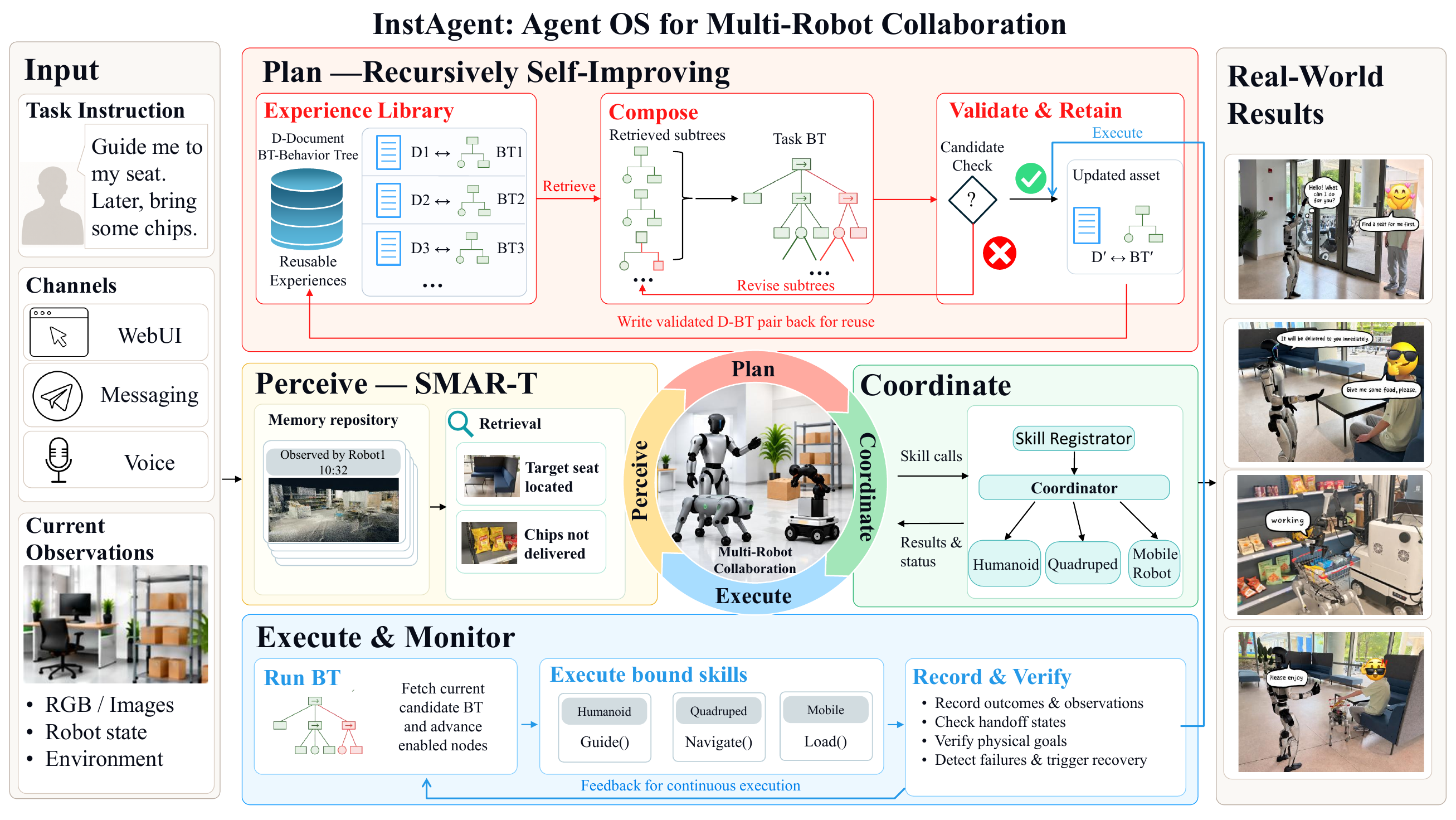}

\caption{
Overview of \textbf{InstAgent}, a recursively self-improving Agent OS for heterogeneous multi-robot collaboration.
InstAgent forms a closed \textbf{Perceive--Plan--Coordinate--Execute} loop: shared observations provide task context, reusable experience is composed into collaborative behavior trees, robot capabilities are coordinated through explicit dependencies and handoffs, and execution outcomes are continuously verified.
Execution feedback revises failed collaboration structures, while validated revisions are retained as reusable document--BT pairs for future inheritance and adaptation.
}
\label{fig:overview}
\end{minipage}\par\vspace{1em}

%% file: sections/00_abstract.tex
\begin{abstract}
Heterogeneous robot teams enable complex, long-horizon tasks by combining complementary sensing, mobility, and manipulation capabilities across multiple embodiments. Yet existing systems struggle to retain execution-time corrections as reusable experience, maintain task-relevant context memory across space and time for reliable handoffs, and explicitly represent the inter-robot dependencies required for structured collaboration. In order to address these challenges, we present \textbf{InstAgent}, a recursively self-improving (RSI) Agent Operating System for heterogeneous robot collaboration. InstAgent organizes team behaviors as agentic behavior trees(BTs) that explicitly model capability assignment, inter-robot dependencies and handoff checks. The RSI mechanism updates these collaboration programs from execution feedback and preserves successful revisions for reuse in future tasks, while shared multimodal spatio-temporal memory provides context for planning and handoff validation. Experiments on 20 real-world multi-robot tasks with three high-level models show consistent gains in task success, first-attempt completion, and execution efficiency.

\par\noindent Project page: \projectpageurl

\end{abstract}

%% file: sections/01_introduction.tex
\section{Introduction}
Heterogeneous robot teams combine complementary capabilities in perception,
mobility and manipulation, enabling long-horizon tasks which are
difficult for a single embodiment to complete. Such tasks, however, require much more than assigning individual skills to different robots. Robots operate
across different locations and stages, and the outcome of one robot may
determine whether another can proceed. Recent large-language-model-based agentic
systems provide a promising abstraction for this setting by connecting
high-level reasoning, heterogeneous skill invocation, multimodal perception,
and execution feedback in a closed loop. This makes them particularly suitable
for long-horizon heterogeneous collaboration, where task procedures cannot
always be fully specified in advance.

Despite this promise, three challenges remain. First, complex physical tasks
rarely succeed from an initial plan alone, yet execution corrections are often
task-specific rather than accumulated as reusable executable experience. Existing
work has explored reflection, skill reuse, and feedback-driven adaptation
~\cite{shinn2023reflexion,wang2023voyager,tziafas2024lifelong,
liu2026phyagentos}, but collaboration procedures corrected during execution
are not always retained in a form that later tasks can inherit and further
improve. Second, long-horizon collaboration requires persistent
spatio-temporal context beyond current local observations, which helps heterogeneous robots plan and act accordingly. Existing memory
systems support spatial and temporal retrieval
~\cite{anwar2025remembr,hu2026raven,zhou2026holoagent,huo2026abot}, but
historical observations must still be connected to current physical checks
before inter-robot handoffs. Third, treating heterogeneous robots as a flat
collection of callable skills under-specifies collaboration structure.
Reliable execution requires explicit inter-robot dependencies, handoff
conditions, and recovery logic, rather than isolated skill invocation alone
~\cite{liu2025coherent,wang2025llm}.

To address the aforementioned challenges, we present \textbf{InstAgent}, a recursively self-improving Agent Operating System designed to enable instant adaptation and deployment of heterogeneous robot teams across diverse and complex scenarios. As shown in Fig.~\ref{fig:overview}, InstAgent represents
collaboration as an executable behavior-tree (BT) program that can be grounded
in shared environmental context, verified during physical execution, and
improved through accumulated experience. Its BT-based recursive
self-improvement (RSI) revises failed collaboration structures from execution
feedback and retains validated revisions for inheritance and further
adaptation. Shared Multimodal Spatio-Temporal Memory with Agent-Driven
Retrieval (\memname{}) provides task-relevant observations with spatial,
temporal, and source information, while current observations verify physical
handoff conditions. InstAgent further composes heterogeneous capabilities into
BTs that explicitly encode inter-robot dependencies, handoff conditions, and
recovery branches. Together, these components form a closed loop of
composition, execution, verification, revision, and reuse.

We evaluate InstAgent on real-world embodiments across 20 heterogeneous
multi-robot tasks and three high-level models. Compared with BT execution without cross-task procedural
experience, feedback-updated experience improves task success by 15--25
percentage points across the three models, while increasing first-attempt
success and reducing recoveries. In the meanwhile, RAVEN-QA independently evaluates task-relevant
environmental observation retrieval.

Our main contributions are:
\begin{itemize}
\item An agentic OS for heterogeneous multi-robot collaboration that
organizes robot capabilities and inter-robot dependencies
into executable BTs for structured coordination and execution.
\item An RSI mechanism that revises collaboration programs from
execution feedback and retains validated revisions for reuse, supported
by shared multimodal spatio-temporal memory (SMAR-T).
\item Extensive real-robot evaluations across heterogeneous platforms,
tasks, and high-level models, demonstrating consistent improvements in
task success and execution performance.
\end{itemize}

%% file: sections/02_related_work.tex
\section{Related Work}

\subsection{Physical Agents and Embodied Execution}

Physical-agent systems connect language reasoning with physical execution
through programs, robot interfaces, and environmental feedback.
Code as Policies~\cite{liang2022codeaspolicies} generates robot programs
that compose perception and control APIs. Inner Monologue~\cite{pmlr-v205-huang23c}
uses execution feedback for closed-loop planning. ROS-LLM~\cite{mower2026robot}
integrates language models with ROS for multiple execution representations,
skill extension, and feedback-driven reflection. RoboClaw~\cite{li2026roboclaw}
uses a VLM to orchestrate learned policy primitives.
Recent AgentOS frameworks further combine spatial memory, heterogeneous
capability scheduling, and runtime management:
HoloAgent-0~\cite{zhou2026holoagent} supports 3D spatial memory and cross-robot execution,
ABot-Claw~\cite{huo2026abot} provides shared cross-embodiment memory and feedback-driven
correction, and ABot-AgentOS~\cite{tian2026abot} stores environmental and task context in a
source-grounded multimodal graph.
These systems support cross-robot coordination and feedback-driven
recovery. Table~\ref{tab:optional-capabilities} summarizes the
capabilities of representative agentic robotic systems, including
collaboration, spatial memory, and the two forms of RSI considered in
this work. Despite these advances, mechanisms for turning
collaboration failures into reusable handoff checks and recovery rules
remain under-specified.
\sys{} combines shared observations with collaboration BTs to revise
inter-robot dependencies and handoff conditions using execution
feedback and retain effective revisions for subsequent tasks.

\input{tables/03_table1}

\subsection{Heterogeneous Multi-Robot Collaboration}

LLM-based multi-robot systems organize joint actions through dialogue
negotiation or hierarchical task allocation.
RoCo~\cite{mandi2024roco} negotiates task plans through inter-robot dialogue and refines
execution using motion planning and environmental
feedback.
Chen et al.~\cite{chen2024scalable} further compare centralized,
decentralized, and hybrid communication architectures in terms of
task success and communication efficiency.
These approaches support flexible and interpretable coordination,
although communication and context requirements increase with team
size and task horizon.
Hierarchical methods separate global coordination from local
execution~\cite{zhang2025lamma}.
COHERENT~\cite{liu2025coherent}, for example, uses a centralized
task assigner to match heterogeneous capabilities and adjusts
plans through feedback from robot executors.

Executing the allocated tasks also requires organizing action
dependencies and recovery procedures.
In robotics, behavior trees (BTs) provide a control structure for
this purpose~\cite{colledanchise2019towards,zeng2026emboteam},
combining reactivity and modularity with ease of modification
and interpretability~\cite{iovino2022survey,cai2025hbtp}.
LLM-HBT~\cite{wang2025llm} integrates this structure with task
allocation, using condition checks, local subtree extension,
and inter-robot task reassignment to recover from failures.
These methods improve coordination and runtime adaptation, but leave
under-specified how corrected handoff conditions and recovery
dependencies are validated, persistently retained, and inherited
for further revision across tasks.
\sys{} retains validated handoff checks and recovery logic as reusable
collaboration subtrees, enabling subsequent tasks to inherit and
further improve effective revisions made during execution.

\subsection{RSI and Executable Experience}

Existing embodied-agent research has explored how execution feedback can
be converted into reusable experience.
Reflexion~\cite{shinn2023reflexion} stores feedback as verbal reflections and represents
Experience-Context RSI, whereas Voyager~\cite{wang2023voyager} and LRLL~\cite{tziafas2024lifelong} convert
experience into retrievable, composable, and executable skills,
representing Structured-Program RSI.
However, these methods primarily target a single agent or single-robot
skills and do not represent role bindings, handoff conditions, or
recovery dependencies in heterogeneous teams.

In the context of physical AgentOS systems, PhyAgentOS~\cite{liu2026phyagentos} organizes
execution, verification, diagnosis, revision, re-verification, and
experience consolidation into a self-evolution loop.
ABot-AgentOS~\cite{tian2026abot} uses failure-driven self-evolution to improve memory writing,
retrieval, and evidence-selection policies.
These systems primarily represent
Experience-Context RSI, rather than directly revising heterogeneous multi-robot
collaboration programs.

Experience-Context RSI is flexible, but translating experience into
concrete execution still depends on the Agent's current reasoning and
does not inherently preserve stable execution orders and explicit
dependencies.
Structured-Program RSI retains conditions, execution orders, and
recovery branches in the program structure, making it better suited to
repeated or structurally similar collaboration tasks.
\sys{} combines both forms of RSI:
Experience-Context RSI uses experience documents, failure feedback, and
shared observations to guide revisions, while Structured-Program RSI
stores validated revisions as collaboration BT subtrees for subsequent
inheritance, rebinding, and improvement.

%% file: tables/03_table1.tex
\begin{table}[t]
\centering
\caption{Capability comparison of agentic robotic systems.\\
$\checkmark$: explicitly reported; $\times$: not explicitly reported.}
\label{tab:optional-capabilities}
\scriptsize
\setlength{\tabcolsep}{2pt}
\renewcommand{\arraystretch}{1.15}

\begin{tabular}{@{}lcccc@{}}
\toprule

\multirow{2}{*}{System}
& \multirow{2}{*}{Collab.}
& \multirow{2}{*}{\shortstack{Spatial\\memory}}
& \multicolumn{2}{c}{RSI} \\

\cmidrule(l){4-5}

& & 
& Experience-Context
& Structured-Program \\

\midrule

LRLL~\cite{tziafas2024lifelong}
& $\times$
& $\times$
& $\checkmark$
& $\checkmark$ \\

LLM-HBT~\cite{wang2025llm}
& $\checkmark$
& $\times$
& $\times$
& $\times$ \\

PhyAgentOS~\cite{liu2026phyagentos}
& $\times$
& $\times$
& $\checkmark$
& $\times$ \\

ABot-AgentOS~\cite{tian2026abot}
& $\times$
& $\checkmark$
& $\checkmark$
& $\times$ \\

HoloAgent-0~\cite{zhou2026holoagent}
& $\checkmark$
& $\checkmark$
& $\checkmark$
& $\times$ \\

ABot-Claw~\cite{huo2026abot}
& $\checkmark$
& $\checkmark$
& $\checkmark$
& $\times$ \\

\midrule

\textbf{\sys{}}
& $\checkmark$
& $\checkmark$
& $\checkmark$
& $\checkmark$ \\

\bottomrule
\end{tabular}
\end{table}

%% file: sections/03_method.tex
\section{InstAgent: A Self-Improving Agent OS for Heterogeneous Collaboration}
\label{sec:method}

\noindent\textbf{System Architecture Overview.}
\sys{} is a ROS~2-based agentic OS that coordinates heterogeneous
robots through BTs. At the lowest level, capabilities are exposed as atomic skills, each wrapped as a minimal Skill BT (see Sec.~\ref{sec:skill_bts}). Guided by task goals,
observations, and retrieved experience, the Agent uses
HeteroComposer (see Sec.~\ref{sec:heterocomposer}) to combine these Skill BTs into a collaboration
BT that specifies cross-robot dependencies, handoff conditions,
and recovery branches. The Coordinator binds its action nodes
to available skills, while the Executive executes the tree
and returns feedback.

Collab-RSI (see Sec.~\ref{sec:rsi}) is the program-level recursive self-improvement
mechanism for collaboration BTs. The Agent uses execution
feedback to revise the tree, which resumes execution after
passing the required checks. After successful execution and
goal verification, the BT is saved as a reusable subtree in
the experience library, with a paired document recording its applicability and invocation conditions.
Later tasks inherit and adapt these subtrees, writing further
validated revisions back for reuse. SMAR-T (see Sec.~\ref{sec:memory}) supports planning
and failure diagnosis by retrieving shared robot observations
according to semantic content, location, and time.


\subsection{Skill BTs: From Atomic Skills to Executable Trees}
\label{sec:skill_bts}

Atomic skills in \sys{} are provided by shared modules and robot modules. Shared modules provide navigation and perception capabilities, while robot modules expose motion, speech interaction, and manipulation capabilities specific to each robot. For executable atomic skills, we represent a concrete skill invocation as a minimal behavior tree containing a single \texttt{Action} node, termed a \emph{Skill BT}, which serves as a basic unit for subsequent task composition. The single-\texttt{Action} tree is the minimal atomic instance. The Skill BTs category in Fig.~\ref{fig:rsi}(a) also includes reusable capability procedures composed from these atomic instances.

To support skill understanding and selection, the multimodal large language model (MLLM) reads functional descriptions, interface definitions, and documented invocation contexts, calling procedures, and usage conditions, and converts them into the common skill representation used by the RSI library. Each experience records the skill's purpose, parameter meanings, and usage constraints, maintaining consistency with the registered interface. The system constructs the minimal Skill BT through a deterministic wrapper: the \texttt{skill} field identifies the interface as \texttt{module/skill}, and \texttt{args} carries the invocation arguments. The resulting tree is stored and linked to its corresponding experience. Before execution, the system checks the BT format, interface existence, and argument consistency with the interface schema. Arguments requiring runtime resolution are checked against the schema once resolved.

During task planning, the Agent selects Skill BTs based on task requirements and skill experiences, and supplies the arguments required by the current task. The selected Skill BTs serve as basic subtrees for constructing the complete task BT; subtree composition and dependencies across robots are detailed in the next subsection.

\subsection{HeteroComposer: Collaborative BT Construction}
\label{sec:heterocomposer}

HeteroComposer selects reusable programs for the current task, with paired documents and their applicability context. Given a task goal $g_k$ and context $x_{k,0}$, the system retrieves candidate entries from the experience document store $D_k$. The MLLM examines the goals and usage scope described in these documents to select applicable experience, then retrieves the associated programs from the subtree store $\mathcal T_k$. The documents and programs jointly form the experience set:
\begin{equation}
\begin{aligned}
\mathcal D_k^{\mathrm{cand}}
&= \operatorname{RetrieveDocs}(D_k; g_k, x_{k,0}),\\
\mathcal D_k^{\mathrm{sel}}
&= \operatorname{Select}_{\mathrm{MLLM}}
(\mathcal D_k^{\mathrm{cand}}; g_k, x_{k,0}),\\
\mathcal E_k
&= \{(d, \mathcal T_k[\ell(d)]) \mid d \in \mathcal D_k^{\mathrm{sel}}\},
\end{aligned}
\label{eq:heterocomposer_retrieval}
\end{equation}
where $\ell(d)$ identifies the subtree associated with document $d$.

The Agent constructs a complete collaborative BT from this experience, prioritizing the reuse of existing program structures and filling uncovered steps with atomic skills. Using the available capability information $\mathcal S_k$, the Agent selects robots or modules for individual actions and specifies task-dependent call arguments. Cross-robot coordination is expressed through sequences, conditions, and recovery branches in the BT, allowing subsequent actions to start once their prerequisite requirements are satisfied.

\subsection{Collab-RSI: Experience-Guided BT Evolution for Heterogeneous Collaboration}
\label{sec:rsi}

Collab-RSI enables program-level recursive self-improvement in \sys{} by revising inherited collaboration procedures and their composition using execution feedback, and retaining validated changes for subsequent tasks. When a collaborative BT constructed by HeteroComposer encounters a failed handoff, an unmet execution dependency, or an insufficient recovery procedure, the Agent identifies the affected coordination steps using the current program, task goal, observations, and related experience. Revisions may modify the internal structure of a reused subtree, such as by adding an outcome check, or modify the relationships among subtrees by adjusting action assignments, handoff conditions, and recovery paths across robots. These revisions address whether the outcome produced by one robot satisfies the conditions required by a subsequent robot. 

\input{figures/01_fig2}

After structural and skill-interface checks, the candidate program is re-executed to verify the relevant handoff conditions and task outcome. When cross-task experience updates are enabled, validated programs are written to the experience library as linked BT structures and experience documents. The BT structures retain inter-robot execution dependencies, handoff checks, and recovery procedures, while the documents record task goals, required capabilities and robot roles, applicability conditions, failure context, and revision rationale. Revisions to existing experience update both the associated structure and its document. A validated program composed from multiple existing subtrees is stored as a new experience entry together with its complete coordination structure and applicable context, making the composition itself retrievable and reusable. Failed execution records are retained as diagnostic context for subsequent analysis and applicability assessment. Let $D_k$ and $\mathcal{T}_k$ denote the document collection and the collection of stored BTs and subtrees in the experience library, respectively. With execution record $\tau_k$ and validation results $v_k$, the persistent update is expressed as
\begin{equation}
(D_{k+1},\mathcal{T}_{k+1})
=
\operatorname{Update}
(D_k,\mathcal{T}_k,\tau_k,v_k).
\label{eq:rsi_update}
\end{equation}

For a subsequent task, the Agent selects applicable documents from the updated Experience library according to the current goal, the recorded applicability conditions, and the available robot capabilities, and retrieves the linked program structures and context. Through the composition process in HeteroComposer, the Agent can inherit previously revised subtrees or use a validated composite program as a reusable unit, preserving its inter-robot dependencies, handoff checks, and recovery logic while adapting it to the current task. New feedback may reveal changed coordination conditions or additional handoff failures, prompting further revisions whose validated results are written back to the same library. A composite program formed in one task thus becomes inherited experience in later tasks and remains available for further reuse, composition, and revision. This closes the recursive loop of inheritance, execution, revision, validation, storage, and subsequent inheritance, accumulating executable experience for collaboration among robots with different capabilities.

\input{algorithms/06_algorithm1}

\subsection{SMAR-T: Shared Multimodal Spatio-Temporal Memory with Agent-Driven Retrieval}\label{sec:memory}

SMAR-T pools robot observations in environmental memory $\mathcal H$ to support collaborative planning and failure diagnosis. These observations complement experience documents $D_k$ and reusable subtrees $\mathcal T_k$: they inform BT construction and revision, while Collab-RSI retains validated programs and their context. Observations from one robot can thus support subsequent tasks involving another.

\textbf{Shared observation memory.}
Each entry records an image $I_i$, its caption $c_i$, visual and caption embeddings $z_i^v,z_i^c$, the observing robot's pose $p_i=(x_i,R_i)$ in the shared map, comprising position $x_i\in\mathbb R^3$ and orientation $R_i\in\mathrm{SO}(3)$, timestamp $t_i$, and source robot $r_i$:
\begin{equation}
\begin{aligned}
c_i   &= G_{\mathrm{cap}}(I_i),\\
z_i^v &= \operatorname{norm}(E_v(I_i)),\\
z_i^c &= \operatorname{norm}(E_t(c_i)),\\
m_i   &= (I_i,c_i,z_i^v,z_i^c,p_i,t_i,r_i).
\end{aligned}
\label{eq:memory_entry}
\end{equation}
Here, $m_i$ denotes one observation, and $G_{\mathrm{cap}}$ uses a vision-language model (VLM) to generate a caption. The visual--text encoding follows standard multimodal practice~\cite{radford2021learning,zhai2023siglip,xue2025improve}. Shared-map poses, timestamps, and source identities link observations across robots and task stages. For example, comparing loading and delivery observations helps diagnose missing-payload failures and informs revisions to handoff checks and recovery branches.

\textbf{Agent-driven retrieval and evaluation loop.}
Figure~\ref{fig:spatiotemporal_memory} illustrates the loop. The Agent identifies semantic, spatial, and temporal constraints, then selects a read-only tool and search domain $\mathcal U\subseteq\mathcal H$. Each call may search the full memory or refine a returned candidate pool, including returning to the full memory on a later call. Tool order depends on the task; the illustrated request establishes the living-room and recent-time scope before retrieving the red-sofa scene.

After each call, the Agent jointly evaluates images, captions, and metadata against the original constraints and updates its task context. Ranking alone does not establish constraint satisfaction. The loop succeeds when evidence supports all explicit constraints and supplies the required response fields; conflicts, missing fields, or unresolved references prompt further retrieval. Each real-robot query allows at most 10 calls across the three tools. If evidence remains insufficient and the pool is empty or the budget is exhausted, the Agent reports insufficient evidence and may request a fresh observation. Sufficient evidence supports a grounded response and, when requested, BT construction
or revision. The Coordinator binds actions to Skills, and
the Executive checks arguments and executes the BT, using
current observations to verify physical conditions.

\textbf{Semantic, spatial, and temporal retrieval tools.}
All tools search $\mathcal U$ and preserve each candidate's image, caption, pose, timestamp, and source. The position tool ranks by $d_p(i\mid\hat x)=\lVert x_i-\hat x\rVert_2$ for query location $\hat x$, using only the position component of $p_i$; orientation is retained for navigation. The time tool ranks by $d_t(i\mid\hat t)=|\Delta t_i-\hat t|$, with observation and query times measured from the same scene origin.

For semantic query $q$, the semantic tool first recalls at most ten frames by visual similarity, then reranks only these frames using their captions:
\begin{equation}
\begin{aligned}
z_q&=\operatorname{norm}(E_t(q)),\\
v_i&=\operatorname{cos}(z_q,z_i^v),\quad
u_i=\operatorname{cos}(z_q,z_i^c),\\
\mathcal C_{10}&=\operatorname{TopK}_{10,\,m_i\in\mathcal U}(v_i),\\
s_i&=(1-\lambda)\widetilde v_i+\lambda\widetilde u_i,\quad
\mathcal C_5=\operatorname{TopK}_{5,\,m_i\in\mathcal C_{10}}(s_i).
\end{aligned}
\label{eq:memory_retrieval}
\end{equation}
The tildes denote min--max normalization within $\mathcal C_{10}$, and $\lambda=0.15$.  The tool returns at most five candidates for Agent evaluation.
\input{figures/02_fig3}

%% file: figures/01_fig2.tex
\begin{figure}[t]
\centering
\includegraphics[width=\columnwidth]{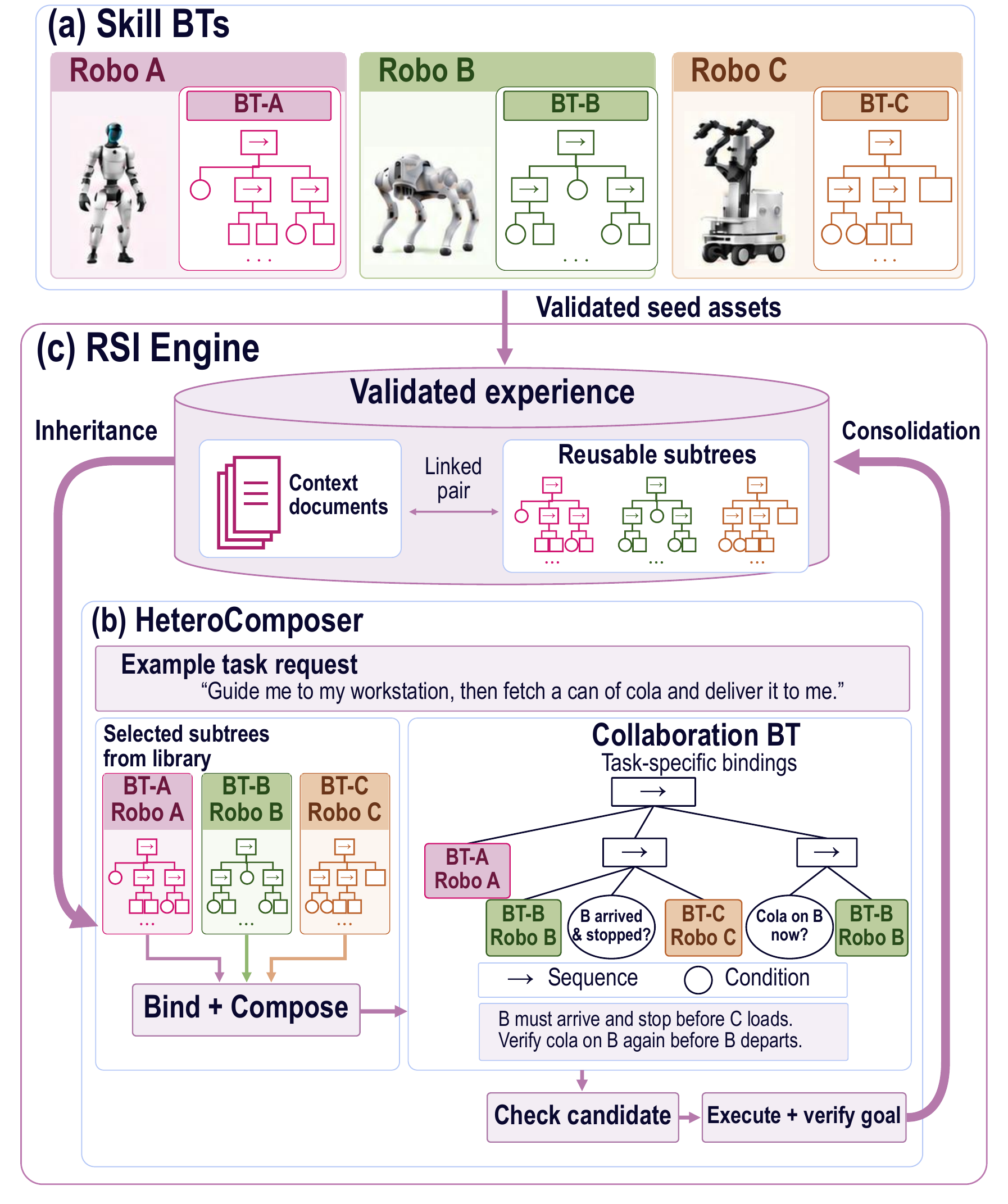}
\caption{BT-based RSI from Skill BTs to collaboration programs. (a) Skill BTs provide reusable capability procedures for task composition. (b) HeteroComposer binds retrieved subtrees to task-specific roles and arguments and composes them with inter-robot handoff conditions. (c) The RSI Engine retains validated programs with linked context documents for inheritance and further revision. BT-A, BT-B, and BT-C denote example Skill BTs associated with robots A, B, and C. The service case in Section~\ref{sec:llm} reports a human-triggered revision followed by automatic reuse.}

\label{fig:rsi}
\end{figure}

%% file: algorithms/06_algorithm1.tex
\begin{algorithm}[t]
\caption{Collab-RSI: Experience-Guided BT Evolution}
\label{alg:rsi}
\footnotesize
\begin{algorithmic}[1]
\Require Goal $g_k$, capabilities $\mathcal S_k$,
         stores $D_k,\mathcal T_k$
\Require Recovery budget $B$, candidate limit $J_{\max}=3$,
         write flag $w$
\Ensure Outcome $o$, trace $\tau_k$,
        stores $D_{k+1},\mathcal T_{k+1}$

\State Observe $x$; $\tau_k\gets\emptyset$;
       $o\gets\textsc{Incomplete}$
\State $\mathcal E_k\gets
       \operatorname{SelectExperience}(D_k,\mathcal T_k;g_k,x)$
\State $T\gets
       \operatorname{ComposeAndCheck}(g_k,x,\mathcal E_k,\mathcal S_k)$
\State \textbf{if} $T$ is unavailable \textbf{then return}
       $o,\tau_k,D_k,\mathcal T_k$

\While{true}
    \State $(o,e,x,\tau_k,v_k,B)\gets$
    \Statex \hspace{\dimexpr\algorithmicindent*2\relax}%
        $\operatorname{ExecuteAndVerify}(T,g_k,\tau_k,B)$
    \State \textbf{if} $o=\textsc{Success}$ or $B=0$
           or $e$ is terminal \textbf{then break}
    \State \textbf{if} safe stopping is unconfirmed
           \textbf{then break}
    \State $T'\gets\operatorname{ReviseAndCheck}_{J_{\max}}
           (T,\mathcal E_k,g_k,x,e)$
    \State \textbf{if} $T'$ is unavailable \textbf{then break}
    \State $T\gets\operatorname{RebuildWithCompletionChecks}(T')$;
           $B\gets B-1$
\EndWhile

\State $(D_{k+1},\mathcal T_{k+1})\gets(D_k,\mathcal T_k)$
\If{$w=1$}
    \State $(D_{k+1},\mathcal T_{k+1})\gets
           \operatorname{Update}(D_k,\mathcal T_k,\tau_k,v_k)$
\EndIf
\State \Return $o,\tau_k,D_{k+1},\mathcal T_{k+1}$
\end{algorithmic}
\end{algorithm}

%% file: figures/02_fig3.tex
\begin{figure*}[t]
\centering
\includegraphics[width=\textwidth]{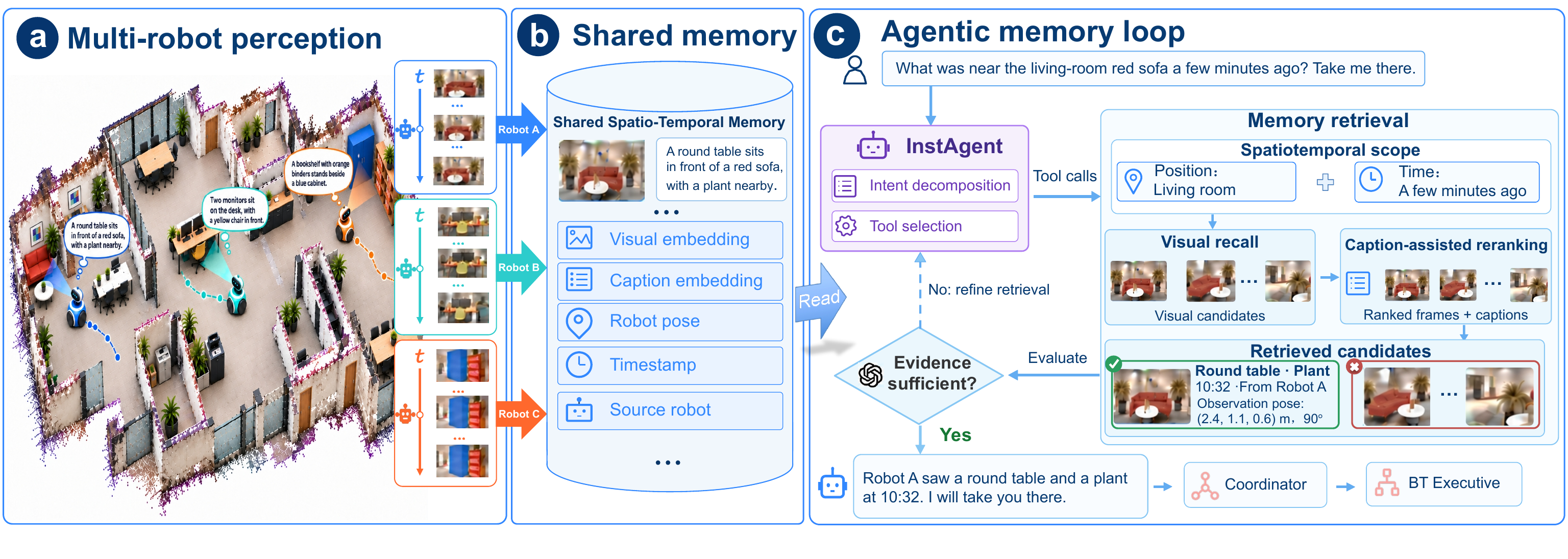}
\caption{SMAR-T shared memory and the InstAgent memory loop. (a) Multiple robots collect observations in a shared map. (b) Each memory entry couples an image and its caption with visual and caption embeddings, the observing robot’s pose, timestamp, and source identity. (c) InstAgent decomposes the request and selects retrieval tools. In the illustrated example, spatial and temporal constraints define the retrieval scope, followed by visual recall and caption-assisted reranking. The Agent evaluates the returned candidates and their metadata against the request. Insufficient evidence triggers further retrieval; sufficient evidence supports a grounded response and, when requested, task execution through the Coordinator and BT Executive.}
\label{fig:spatiotemporal_memory}
\end{figure*}

%% file: sections/04_experiments.tex
\section{Experiments}\label{sec:experiments}

We evaluate BT-based execution, revision, and procedural-experience
reuse through real-robot comparisons and a service case, and assess
task-relevant environmental context through RAVEN-QA retrieval.

\subsection{Experimental Setup}
\label{sec:setup}

\noindent\textbf{Platforms, tasks, and paired trials.}
G1, Go2, and ALOHA provide interaction and guidance, navigation
and transport, and manipulation and loading, respectively;
shared navigation and perception modules return state and execution
feedback. We evaluate 20 heterogeneous multi-robot tasks involving
capability allocation and inter-robot handoffs. Recurring collaboration
structures vary in objects, locations, and goals, testing subtree reuse
and adaptation within a fixed robot team under the perturbations below.
Each of three models runs every task twice under four configurations:
40 runs per model--configuration and 480 measured runs in total.
For each model, corresponding task repetitions are paired across
configurations, with task instances, failure settings, and perturbation
types, injection counts, and trigger stages held constant.

\noindent\textbf{Success criteria and recovery budgets.}
Perturbations include excessive arrival error, navigation and loading
failures, and user-requested destination changes. Relative movement
and turning must be within 0.10\,m and $10^\circ$; workstation navigation
requires a successful module return and human confirmation of arrival
within 0.5\,m. Success requires all physical and interaction
postconditions to be verified. Each task permits at most five executed
failure-triggered recoveries or retries; a goal change alone consumes
no retry. For BT conditions, each failure episode additionally permits
three candidate revisions, with the counter reset per episode.
Rejected candidates consume candidate attempts, not executed-recovery
budget. An incomplete run fails when the recovery budget is exhausted
or, if BT revision is required, all three candidates are rejected or
safe stopping cannot be confirmed (Algorithm~\ref{alg:rsi}).

\noindent\textbf{Configurations and experience control.}
\textbf{Simple CoT} invokes Skills from a textual plan without a BT,
under the same tasks, success criteria, and recovery budget.
\textbf{BT Recovery ($E_0$)} builds BTs from registered atomic Skills
without cross-task subtree retrieval.
\textbf{Fixed-Experience BT ($E_1$)} additionally retrieves existing
subtrees while preserving their executable versions.
\textbf{RSI-Updated Experience BT ($E_2$, full \sys{})} starts from
the same base subtrees as $E_1$, but uses matched historical feedback
and case availability to revise, validate, and retain them before
evaluation.

All BT conditions allow the same failure recovery, task-local
structural revision, and retries under identical budgets. Each model
has its own procedural library; the $E_1$ and $E_2$ snapshots are
prepared before evaluation and frozen during measured runs.
Task-local repairs are not written back, so one run cannot update
another's library. Evaluation tasks are excluded from library
construction. Tasks, models, prompts, perturbations, and recovery
budgets are shared across BT conditions; $E_1$ and $E_2$ also use
the same retrieval procedure. Thus, $E_0$--$E_1$ tests experience
reuse, whereas $E_1$--$E_2$ tests retention of validated revisions.
Model parameters, atomic Skills, low-level controllers, VLA/VLM
components, and task specifications remain fixed.

\noindent\textbf{Metrics.}
Task success rate (SR) measures complete task success.
First-attempt success rate (FSR) requires success without
failure-triggered recovery or retry; a user-requested goal change
alone does not count as a retry. Initial-BT logical coherence (LC)
follows the precondition-respecting execution-order criterion
of~\cite{ao2025llm} for action, stage, and handoff dependencies.
Alternative valid orders are accepted, repaired trees are excluded,
and LC does not apply to Simple CoT. Mean recovery count
($\bar R$) averages failure-triggered recoveries and retries
over all runs, including ultimately unsuccessful ones.

\noindent\textbf{RAVEN-QA evaluation protocol.}
RAVEN-QA~\cite{hu2026raven} provides sparse observations with timestamps
and poses from RealWorld (robot exploration in laboratories and public
areas), Habitat-Sim (simulated robot-view trajectories), IRS (indoor/outdoor
web videos), and IRS-hard (challenging self-recorded/web egocentric videos).
Queries cover object retrieval, visual and commonsense reasoning,
information recall, and spatial understanding. We run RAVEN~\cite{hu2026raven},
ReMEmbR~\cite{anwar2025remembr}, and both InstAgent variants under the
official RAVEN-QA protocol, using identical splits, queries,
preprocessing, QQMM-v2 embeddings~\cite{xue2025improve}, success
criteria, and evaluation scripts. RAVEN and ReMEmbR retain their
respective retrieval protocols; model configurations appear in
Table~\ref{tab:memory_results}. Query-level SR is the percentage of
queries with the requested observation or associated answer correctly
identified. Full \sys{} uses Gemini-2.5-Flash for tool selection,
evidence interpretation, and answer generation.
\emph{Ours w/o Agent} retains the same indexed observations, captions,
embeddings, visual recall, and caption reranking, but returns Top-1
directly. The full system adds candidate comparison, position or time
retrieval when needed, and caption verification against source images.

\subsection{Model Comparison and BT-Based RSI Evaluation}
\label{sec:llm}

\input{tables/04_table2}

\noindent\textbf{Execution and experience.}
Table~\ref{tab:llm} compares three high-level models using
Simple CoT and three BT experience stages.
Baseline planning and atomic Skills can complete some tasks.
Across all three models, Full \sys{} achieves higher SR and
FSR and fewer recoveries than BT Recovery, supporting the
benefit of retained and updated procedural experience on the
20 evaluated tasks.

\noindent\textbf{Experience-stage progression.}
Table~\ref{tab:llm} shows performance across fixed
experience stages. The primary contrast is $E_0$--$E_2$: across the
three models, feedback-updated experience improves SR by 15, 22.5,
and 25 points, FSR by 12.5, 30, and 20 points, and reduces
$\bar R$ by 0.70, 1.30, and 1.20. We compare $E_0$ and $E_2$
at the task-instance level using McNemar's test for paired
task-success outcomes and the Wilcoxon signed-rank test for
paired recovery counts. Both tests yield $p<0.05$ for all
three models.
The $E_0$--$E_1$ comparison evaluates the benefit of reusing
existing procedural experience. The $E_0$--$E_2$ contrast measures
the combined benefit of experience reuse and retained revisions.
With base subtrees, historical feedback, and case availability
matched, $E_1$--$E_2$ assesses the additional benefit associated
with retaining validated revisions in the executable subtrees.
These comparisons assess program reuse and adaptation across object, location, and goal variations using prepared library snapshots.
Relative to $E_1$, the observed SR gains of $E_2$ are
10, 17.5, and 15 percentage points for GPT-5.6 Sol,
DeepSeek-V4-Pro, and Qwen3.8-27B, respectively.
LC also increases monotonically from $E_0$ to $E_2$
across all three models.

\noindent\textbf{Collaboration-program revision and reuse.}
This qualitative case, excluded from the aggregate results, illustrates
handoff diagnosis, program revision, and subsequent reuse.
After G1 guides the user to a workstation, Go2 collects cola from ALOHA
and returns after loading verification. Shared item, loading-location,
and destination arguments connect the capability subtrees.

In one run, cola was manually removed after loading confirmation but
before Go2 departed. After arrival, a tester's web-interface report of
missing cola prompted the Agent to retrieve historical observations and
identify its absence during transport. Reflection and replanning added
VLM checks of current basket contents by ALOHA before return navigation
and G1 before delivery; either failed check triggers reloading.
The validated subtree retains both checks and recovery logic for
automatic reuse, while its linked document records the diagnosis and
revision rationale (Fig.~\ref{fig:rsi}). Atomic Skills remain unchanged.

\subsection{RAVEN-QA Memory Evaluation}
\label{sec:memory_eval}

\input{tables/08_memory_results}

The methods differ in how they represent and select visual
experience. RAVEN directly indexes original observations with
multimodal visual embeddings, preserving image details but using no
language description as a second ranking signal~\cite{hu2026raven}.
ReMEmbR retrieves caption-based memories~\cite{anwar2025remembr},
making semantics explicit but losing details omitted by captioning.
Our method combines their complementary signals: visual similarity
first forms a high-recall candidate set, captions refine its order,
and the Agent verifies the selected captions against their original
images. Language thus assists selection without replacing visual
evidence.

Our method reaches 93.14\% overall, exceeding RAVEN by 8.57 points
and ReMEmbR by 28.65 points. Its largest gains over RAVEN occur on
RealWorld (6.19 points) and Habitat-Sim (13.32 points), consistent
with caption reranking helping distinguish visually similar scenes
by object identity, attributes, or role. IRS leaves only a
1.85-point margin because direct visual retrieval is already strong.
The near tie on IRS-hard (82.93\% versus 82.90\%) exposes the
method's boundary: reranking cannot recover targets that are absent,
occluded, or missed by the visual candidate set.

Returning Top-1 without the Agent reduces overall SR from 93.14\%
to 69.31\%, with the largest drops on Habitat-Sim and IRS-hard.
The Agent is a unified retrieval controller that performs candidate
comparison, position and time tool calls, evidence interpretation,
and caption verification. Thus, this 23.83-percentage-point gap
measures the overall contribution of the integrated Agent-driven
retrieval workflow rather than the contribution of any individual
component. The smaller IRS drop is consistent with its more direct
queries, for which Top-1 is more often sufficient.

%% file: tables/04_table2.tex
\begin{table}[t]
\centering
\caption{Real-robot performance across high-level models,
execution configurations, and procedural-experience stages.}
\label{tab:llm}
\footnotesize
\setlength{\tabcolsep}{2.3pt}
\renewcommand{\arraystretch}{1.08}

\begin{tabular*}{\columnwidth}{@{\extracolsep{\fill}}lcccc@{}}
\toprule
Configuration
& SR$\uparrow$
& FSR$\uparrow$
& LC$\uparrow$
& $\bar R\downarrow$ \\
\midrule

\multicolumn{5}{@{}l}{\textbf{GPT-5.6 Sol}} \\
Simple CoT
& 25/40 & 20/40 & --- & 1.60 \\
BT Recovery ($E_0$)
& 32/40 & 30/40 & 35/40 & 1.10 \\
Fixed-Experience BT ($E_1$)
& 34/40 & 32/40 & 37/40 & 0.85 \\
Full \sys{} ($E_2$)
& 38/40 & 35/40 & 39/40 & 0.40 \\
\midrule

\multicolumn{5}{@{}l}{\textbf{DeepSeek-V4-Pro}} \\
Simple CoT
& 22/40 & 14/40 & --- & 2.15 \\
BT Recovery ($E_0$)
& 29/40 & 24/40 & 33/40 & 1.65 \\
Fixed-Experience BT ($E_1$)
& 31/40 & 28/40 & 35/40 & 1.30 \\
Full \sys{} ($E_2$)
& 38/40 & 36/40 & 39/40 & 0.35 \\
\midrule

\multicolumn{5}{@{}l}{\textbf{Qwen3.8-27B}} \\
Simple CoT
& 17/40 & 11/40 & --- & 2.70 \\
BT Recovery ($E_0$)
& 24/40 & 21/40 & 28/40 & 2.20 \\
Fixed-Experience BT ($E_1$)
& 28/40 & 24/40 & 32/40 & 1.80 \\
Full \sys{} ($E_2$)
& 34/40 & 29/40 & 36/40 & 1.00 \\

\bottomrule
\end{tabular*}

\par\smallskip
\begin{minipage}{\columnwidth}
\footnotesize
\raggedright
Each model--configuration pair includes 40 real-robot runs
(20 tasks, two runs per task).
SR, FSR, and LC are counts out of 40; $\bar R$ averages
failure-triggered recoveries and retries over all runs.
All BT conditions allow task-local revision under the same
recovery budget.
$E_1$ and $E_2$ share historical feedback and cases; only $E_2$
persists validated revisions to inherited subtrees before evaluation.
Both procedural libraries remain frozen during evaluation.
LC evaluates initial BTs and does not apply to Simple CoT (---).
\end{minipage}
\end{table}

%% file: tables/08_memory_results.tex
\begin{table}[t]
\centering
\caption{Query-level task success rate (SR, \%) on RAVEN-QA.
Higher is better.}
\label{tab:memory_results}
\footnotesize
\setlength{\tabcolsep}{2.7pt}
\renewcommand{\arraystretch}{1.1}

\begin{tabular}{@{}lrrrr@{}}
\toprule
Subset & RAVEN & ReMEmbR & Ours & \shortstack{Ours w/o\\Agent} \\
\midrule
RealWorld   & 89.05 & 76.67 & \textbf{95.24} & 76.19 \\
Habitat-Sim & 80.47 & 54.99 & \textbf{93.79} & 63.35 \\
IRS         & 96.30 & 92.60 & \textbf{98.15} & 92.59 \\
IRS-hard    & 82.90 & 58.50 & \textbf{82.93} & 58.54 \\
\midrule
Overall     & 84.57 & 64.49 & \textbf{93.14} & 69.31 \\
\bottomrule
\end{tabular}

\par\smallskip
\begin{minipage}{\columnwidth}
\scriptsize
All results are from our own runs.
RAVEN, ReMEmbR, and Ours use QQMM-v2 and Gemini-2.5-Flash.
Ours w/o Agent retains the same embedder and directly returns
the Top-1 retrieval result without a VLM Agent.
\end{minipage}
\end{table}

%% file: sections/05_conclusion.tex
\section{Conclusion}
\sys{} connects capability binding, shared context, and physical execution through an agent operating system with BT-based RSI. Linked documents and executable subtrees retain feedback-driven changes to collaboration programs, allowing inherited procedures to be composed and revised again. Real-robot evaluation across three models shows higher task and first-attempt success with feedback-updated experience than with cold starts on the evaluated tasks. The service case illustrates how a retained BT condition changes the handoff between loading and transport. Future work will examine transfer to new task compositions and robot teams, and the reliability of experience updates over longer deployment periods.

%% file: sections/06_acknowledgments.tex
\section*{Acknowledgments}
This work was supported by the National Natural Science Foundation of China (62406195, 62303319), the HPC Platform of ShanghaiTech University, the Core Facility Platform of Computer Science and Communication of ShanghaiTech University, the Key Laboratory of Intelligent Perception and Human-Machine Collaboration (ShanghaiTech University), Ministry of Education, and the Shanghai Engineering Research Center of Intelligent Vision and Imaging.